\documentclass{article}
\usepackage{ijcai22}

\usepackage{times}
\usepackage{soul}
\usepackage{url}
\usepackage[hidelinks]{hyperref}
\usepackage[utf8]{inputenc}
\usepackage[small]{caption}
\usepackage{graphicx}
\usepackage{multicol}
\usepackage{multirow}
\usepackage{amsmath}
\usepackage{amsfonts}
\usepackage{amsthm}
\usepackage{booktabs}
\usepackage{algorithm}
\usepackage{algorithmic}
\usepackage[ruled,vlined,algo2e]{algorithm2e}
\usepackage{booktabs}
\usepackage{color}
\usepackage{xcolor}
\usepackage{mathtools}
\usepackage[normalem]{ulem}
\usepackage{stmaryrd}

\newtheorem{theorem}{Theorem}

\newtheorem{definition}{Definition}

\title{Computing Rule-Based Explanations of Machine Learning Classifiers using Knowledge Graphs}

\author{
Edmund Dervakos
\and
Orfeas Menis - Mastromichalakis\and
Alexandros Chortaras\And
Giorgos Stamou
\affiliations
Artificial Intelligence and Learning Systems Laboratory \\ 
School of Electrical and Computer Engineering \\
National Technical University of Athens
\emails
menorf@ails.ece.ntua.gr,
eddiedervakos@islab.ntua.gr,
\{achort,gstam\}@cs.ntua.gr 
}

\begin{document}

\maketitle

\begin{abstract}

The use of symbolic knowledge representation and reasoning as a way to resolve the lack of transparency of machine learning classifiers is a research area that lately attracts many researchers. In this work, we use knowledge graphs as the underlying framework providing the terminology for representing explanations for the operation of a machine learning classifier. In particular, given a description of the application domain of the classifier in the form of a knowledge graph, we introduce a novel method for extracting and representing black-box explanations of its operation, in the form of first-order logic rules expressed in the terminology of the knowledge graph.

\end{abstract}

\section{Introduction}

Machine learning systems' explanations need to be represented in a human-understandable form, employing the standard domain terminology and this is why symbolic AI systems play a key role in the eXplainable AI (XAI) field of research   \cite{DBLP:journals/corr/abs-1901-04592,DBLP:journals/csur/GuidottiMRTGP19,DBLP:journals/inffus/ArrietaRSBTBGGM20}.
Of great importance in the field are the so called \emph{rule-based explanation} methods.  
Many of them rely on statistics to generate lists of if-then rules which mimic the behaviour of a classifier \cite{DBLP:conf/icml/YangRS17,DBLP:journals/tvcg/MingQB19}, or extract rules in the form of decision trees \cite{DBLP:conf/nips/CravenS95,DBLP:journals/corr/abs-1906-08362}, while some methods make use of logics \cite{DBLP:journals/apin/LehmannBH10,DBLP:conf/nesy/SarkerXDRH17} and extract rules in a form that it can be argued to be the desirable form of explanations \cite{DBLP:conf/aaai/PedreschiGGMRT19}. Concerning the vocabulary they use, most approaches generate rules in terms of the feature space of the black-box classifier. However, it is argued that when the feature space of the classifier is sub-symbolic raw data, providing explanations in terms of features might lead to unintuitive, or even misleading results \cite{mittelstadt2019explaining}. Some recent methods utilize additional information about the data (such as objects depicted in an image) \cite{DBLP:conf/ijcai/CiravegnaGGMM20}, or external semantic information for the data \cite{panigutti2020doctor}.
The requirement for expressing explanations in terms of domain knowledge with formal semantics has motivated the use of knowledge graphs (KG) \cite{DBLP:journals/corr/abs-2003-02320} in XAI \cite{tiddi2021knowledge}. Knowledge graphs allow for the development of mutually agreed upon terminology to describe a domain in a human-understandable and computer-readable manner. In this respect, knowledge graphs have emerged as a promising complement or extension to machine learning approaches for explainability \cite{xaisurvey}, like explainable recommender systems \cite{ai2018learning} that make use of knowledge representation, explainable Natural Language Processing pipelines \cite{silva2019exploring} which utilize knowledge graphs such as WordNet, and computer vision approaches, explainable by incorporating external knowledge \cite{Alirezaie2018ASA}. 

Following this line of work, we approach the problem of explaining the operation of opaque, black-box deep learning classifiers as follows: a) using domain knowledge, we construct a set of characteristic semantically-described items in the form of a knowledge graph (which we call an \emph{explanation dataset}), b) we check the output of the unknown classifier against these items, and c) we describe the common characteristics of class instances in a human-understandable form (as if-then rules), following a reverse engineering procedure. For the latter, we propose a novel method for automatically extracting and representing global, post hoc explanations as first-order logic expressions produced through \emph{semantic queries} over the knowledge graph, covering interesting common properties of the class instances. In this way, the problem of extracting logical rules is approached as a \emph{semantic query reverse engineering problem}. Specifically, in order to extract rules of the form ``if an image depicts X then it is classified as Y'', we acquire the set of items classified as ``Y'' and then we reverse engineer a semantic query bound to have this set of items as certain answers. The use of semantic queries in our framework allows us to utilize the strong theoretical and practical results in the area of semantic query answering  \cite{DBLP:journals/jar/CalvaneseGLLR07,DBLP:journals/kais/TrivelaSCS20}. The query reverse engineering problem has been studied in the context of databases \cite{DBLP:journals/vldb/TranCP14} and more recently of SPARQL queries \cite{DBLP:conf/www/ArenasDK16}. Here, we use expressive knowledge graphs, where the answers to queries are considered to be \emph{certain answers}, whose computation involves reasoning.

The theoretical results in Sec.~\ref{sec:Rule-based global explanations} and \ref{sec:Semantic queries as explanation rules} show that we can provide the user with formal guarantees for the  extracted rule-based explanations. The experiments and comparative evaluation in Sec. \ref{sec-evaluation} show that the results of our approach are promising, performing similarly with state-of-the-art rule-based explainers in settings where the classifier is feature based, while showing a clear improvement in more realistic settings of deep learning classifiers (with raw data like images as input), especially in the presence of expressive knowledge. 

\section{Background}
\label{sec:preliminaries}

\textbf{Description Logics} Let $\mathcal{V}=\langle \mathsf{CN},\mathsf{RN},\mathsf{IN}\rangle$ be a \emph{vocabulary}, where $\mathsf{CN}$, $\mathsf{RN}$, $\mathsf{IN}$ are mutually disjoint finite sets of \emph{concept}, \emph{role} and \emph{individual} names, respectively. Let also  $\mathcal{T}$ and $\mathcal{A}$ be a terminology (TBox) and an assertional database (ABox), respectively, over $\mathcal{V}$ using a Description Logics (DL) dialect $\mathcal{L}$, i.e. a set of axioms and assertions that use elements of $\mathcal{V}$ and constructors of $\mathcal{L}$. The pair $\langle\mathcal{V},\mathcal{L}\rangle$ is a \emph{DL-language}, and $\mathcal{K}=\langle\mathcal{T},\mathcal{A}\rangle$ is a \emph{(DL) knowledge base} (KB) over this language. 
The semantics of KBs are defined the standard model theoretical way using interpretations. 
Given a non-empty domain $\Delta$, an interpretation $\mathcal{I}=(\Delta^{\mathcal{I}},\cdot^{\mathcal{I}})$ assigns a set  $C^{\mathcal{I}} \subseteq \Delta^{\mathcal{I}}$ to each  $C\in\mathsf{CN}$, a set $r^{\mathcal{I}} \subseteq \Delta^{\mathcal{I}} \times \Delta^{\mathcal{I}}$ to each  $r\in\mathsf{RN}$, and an $a^{\mathcal{I}}\in\Delta$ to each  $a\in\mathsf{IN}$.
$\mathcal{I}$ is a \emph{model} of a KB $\mathcal{K}$ iff it satisfies all assertions in $\mathcal{A}$ and all axioms in $\mathcal{T}$. 

\noindent \textbf{Conjunctive queries} A \emph{conjunctive query} (simply, a \emph{query}) $q$ over a vocabulary $\mathcal{V}$ is an expression $\{\,\langle x_1,\ldots x_k\rangle \mid \exists y_1\ldots\exists y_l.(c_1\wedge\ldots\wedge c_n)\,\}$, where $k,l\geq 0$, $n\geq 1$, $x_i,y_i$ are  variables, each $c_i$ is an atom $C(u)$ or $r(u,v)$, where $C \in \mathsf{CN}$, $r \in \mathsf{RN}$, $u,v$ are some $x_i$, $y_i$ or in $\mathsf{IN}$, and all $x_i,y_i$ appear in at least one atom. The vector $\langle x_1,\ldots x_k\rangle$ is the \emph{head} of $q$, its elements are the \emph{answer variables}, and $\{c_1,\ldots, c_n\}$ is the \emph{body} of $q$. For simplicity, we write queries as $q\doteq\{c_1,...,c_n\}_{x_1,\ldots,x_k}$.  
$\mathsf{vars}(q)$ is the set of all variables appearing in $q$.
A query $q$ can also be viewed as a graph, with a node $v$ for each element in  $\mathsf{vars}(q)$ and an edge $(u,v)$ if there is an atom $r(u,v)$ in  $q$, and labeling nodes and edges by the respective atom predicates. A query is \emph{connected} if its graph is connected.
In this paper we focus on connected queries having \emph{one} answer variable in which all arguments of all $c_i$s are variables, which we call \emph{instance queries}. 
A query $q_2$ \emph{subsumes} a query $q_1$ (we write $q_1 \leq_S q_2$) 
iff there is a substitution  $\theta$ s.t. $q_2\theta \subseteq q_1$. If $q_1$, $q_2$ are mutually subsumed, they are \emph{syntactically equivalent}. 
Let be $q$ a query and $q'\subseteq q$. If $q'$ is a minimal subset of $q$ s.t. $q'\leq_S q$, then $q'$ is a \emph{condensation} of $q$ ($\mathsf{cond}(q)$).  
Given a KB $\mathcal{K}$, an instance query $q$ and an interpretation $\mathcal{I}$ of $\mathcal{K}$, a \emph{match} for $q$ is a mapping $\pi:\mathsf{vars}(q)\rightarrow \Delta^\mathcal{I}$ such that $\pi(u)\in C^\mathcal{I}$ for all $C(u) \in q$, and $(\pi(u),\pi(v)) \in r^\mathcal{I}$ for all $r(u,v) \in q$.
Then, $a$ is a \emph{(certain) answer} for $q$ over $\mathcal{K}$ if in every model $\mathcal{I}$ of $\mathcal{K}$ there is a match $\pi$ for $q$ such that $\pi(x)=a^\mathcal{I}$. We denote the set of certain answers (\emph{answer set}) to $q$ by $\mathsf{cert}(q,\mathcal{K})$. Because computing $\mathsf{cert}(q,\mathcal{K})$ involves reasoning on $\mathcal{K}$, queries on top of DL KBs are characterized as \emph{semantic queries}.

\noindent \textbf{Rules} A (definite Horn) \emph{rule} is a FOL expression of the form $\forall x_1\ldots \forall x_n\, ( c_1,\ldots, c_n \Rightarrow c_0)$, usually written as $ c_1,\ldots,c_n\rightarrow c_0$,  where the $c_i$s are atoms and $x_i$ all appearing variables. In a rule over a vocabulary $\mathcal{V}$, each $c_i$ is either $C(u)$ or $r(u,v)$, where $C\in\mathsf{CN}$, $r\in\mathsf{RN}$. The body of a rule can be represented as a graph similarly to the queries. In this paper we assume that all rules are \emph{connected}, i.e. the graph of the body extended with the head variables is connected.

\noindent \textbf{Classifiers} A classifier is viewed as a function $F:\mathcal{D}\rightarrow{\mathcal{C}}$, where $\mathcal{D}$ is a domain of item feature data (e.g. images, audio, text), and $\mathcal{C}$ a set of classes (e.g. $\mathsf{Dog}$, $\mathsf{Cat}$). 

\section{Rule-based global explanations}
\label{sec:Rule-based global explanations}

\begin{figure}[h!]
    \centering
    \includegraphics[width=1.0\linewidth]{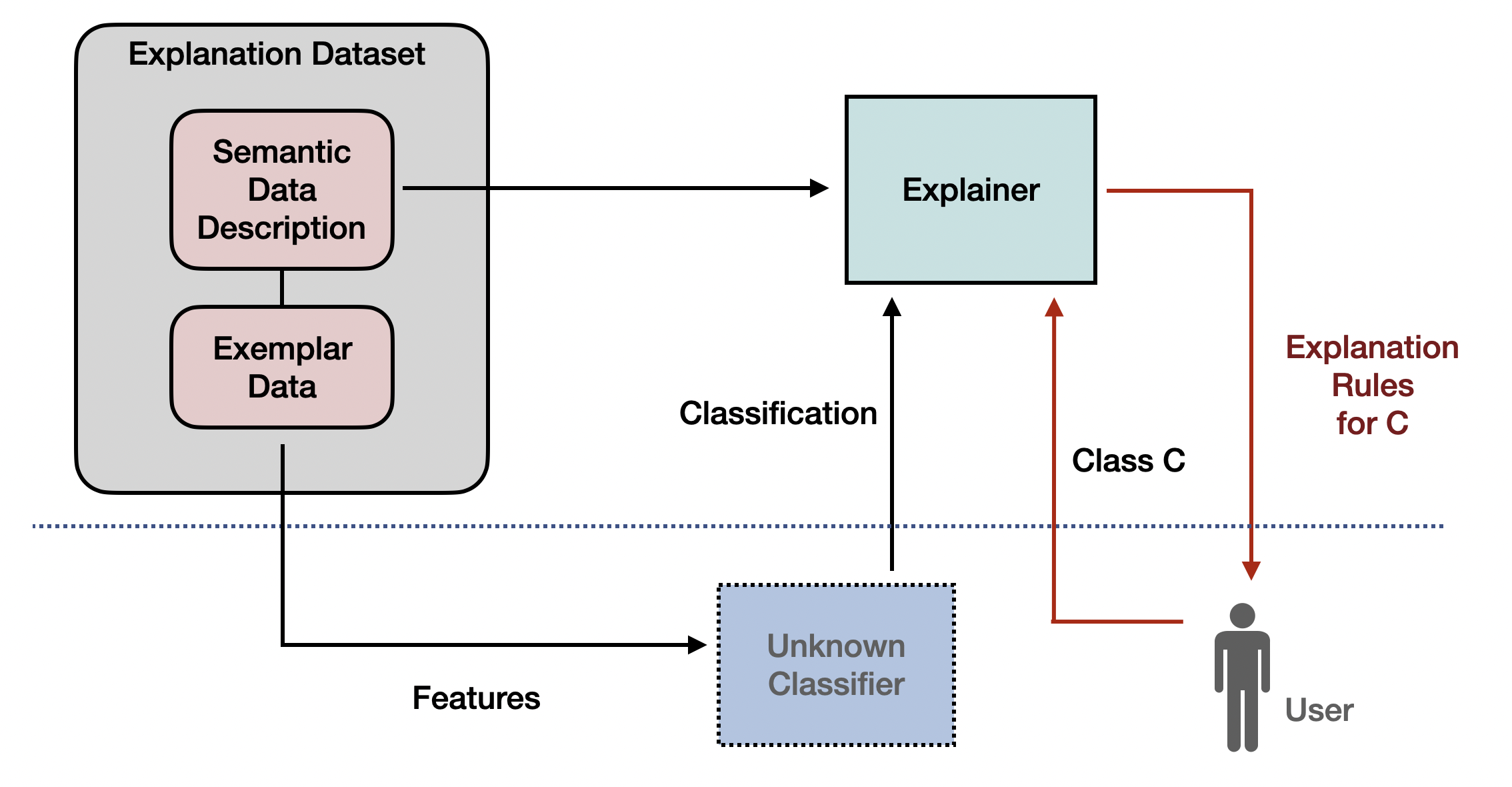}\vspace*{-4pt}
    \caption{Explainer in operation}
    \label{fig:general_architecture}
\end{figure}

Our approach to the extraction of rule-based global explanations is shown in Fig. \ref{fig:general_architecture}. The \emph{explainer} takes as input the output of an \emph{unknown classifier} to specific items (the \emph{exemplar data}) and a \emph{class $C$} from the user, and computes \emph{explanation rules for $C$}, in the form of definite Horn rules. The  explanation rules are expressed using a standard vocabulary  (e.g. terms from domain ontologies), which is understandable and useful to the end-user. 
To compute the explanation rules, the explainer has access also to \emph{semantic data descriptions} associated to the exemplar data items, expressed in the same vocabulary. The exemplar data, that are the items fed to the unknown classifier, and their associated semantic data descriptions comprise an \emph{explanation dataset}. 

In this paper we consider semantic data descriptions that are expressed as DL knowledge bases, and in order to compute the explanation rules with \emph{semantic guarantees} we use semantic query answering technologies, taking advantage of the semantic interrelation of rules and conjunctive queries over DL knowledge bases \cite{DBLP:journals/jacm/MotikR10}. Intuitively, given a class $C$ and using semantic query answering, the explainer computes and expresses as rules the conjunctive queries that have as answers individuals representing the exemplar data items that the unknown classifier classifies as~$C$.

Because the exemplar data are consumed by the classifier,  
we consider that each exemplar data item consists of all the information that the classifier needs to classify it (the necessary \emph{features}). The association of a semantic data description to each such item is modeled by the explanation dataset.

\begin{definition}[Explanation Dataset] \label{def:explanation dataset}

Let $\mathcal{D}$ be a domain of item feature data, $\mathcal{C}$ a set of classes, and $\mathcal{V}=\langle{\mathsf{IN, CN, RN}}\rangle$ a vocabulary such that $\mathcal{C}\cup\{\mathsf{Exemplar}\} \subseteq \mathsf{CN}$. Let also $\mathsf{EN} \subseteq \mathsf{IN}$ be a set of \emph{exemplars}.
An \emph{explanation dataset} $\mathcal{E}$ in terms of $\mathcal{D}$, $\mathcal{C}$, $\mathcal{V}$ is a tuple $\mathcal{E}= \langle{ \mathcal{M},\mathcal{S}}\rangle$, where $\mathcal{M}:{\mathsf{EN}} \rightarrow \mathcal{D}$ is a mapping from the exemplars to the item feature data, and $\mathcal{S}=\langle\mathcal{T},\mathcal{A}\rangle$ is a DL KB over $\mathcal{V}$ such that $\mathsf{Exemplar}(a)\in \mathcal{A}$ iff $a\in\mathsf{EN}$, the elements of $\mathcal{C}$ do not appear in $\mathcal{S}$, and $\mathsf{Exemplar}$ and the elements of $\mathsf{EN}$ do not appear in~$\mathcal{T}$.
\end{definition}

Intuitively, $\mathcal{D}$ contains the items, as feature data, that  can be fed to a classifier. Each such item is represented in the associated semantic data description by an individual (exemplar) $a\in\mathsf{EN}$, which is mapped to the respective feature data by $\mathcal{M}$. $\mathcal{S}$ contains the semantic data descriptions about all individuals in $\mathsf{EN}$. The concept $\mathsf{Exemplar}$ is used solely to identify the exemplars within $\mathcal{A}$ (since other individual may exist) and should not appear elsewhere. The classes $\mathcal{C}$ should not appear in $\mathcal{S}$ so as not to take part in any reasoning process.

Given an explanation dataset, an unknown classifier, and a class $C$, the aim of the explainer is to detect the semantic properties and relations of the  exemplar data items that are classified by the unknown classifier to class $C$, and represent them in a human-understandable form, as rules.

\begin{definition}[Explanation Rule] \label{def:explanation rule}

Let $F:\mathcal{D}\rightarrow{\mathcal{C}}$ be a classifier
, $\mathcal{E}=\langle{ \mathcal{M},\mathcal{S}}\rangle$ an explanation dataset in terms of $\mathcal{D}$, $\mathcal{C}$ and an appropriate vocabulary $\mathcal{V}=\langle{\mathsf{CN, RN, IN}}\rangle$. Given a concept $C \in \mathcal{C}$, the rule 
\[
\mathsf{Exemplar}(x), c_1,c_2,\ldots,c_n\rightarrow C(x)
\]
\noindent where $c_i$ is an atom $D(u)$ or $r(u,v)$, where $D \in \mathsf{CN}$, $r \in \mathsf{RN}$, and $u,v$ are variables, is an \emph{explanation rule} of $F$ for class $C$ over  $\mathcal{E}$. 
We denote the rule by $\rho(F,\mathcal{E},C)$, or simply by $\rho$ whenever the context 
is clear.
We may also omit $\mathsf{Exemplar}(x)$ from the body, since it is a conjunct of any explanation rule.
\end{definition}

Explanation rules describe \textit{sufficient} conditions for an item  to be classified in class $C$ by a classifier. E.g., if the classifier classified images depicting wild animals in a zoo class, an explanation rule could be $\mathsf{Exemplar}(x),\mathsf{Image}(x),\mathsf{depicts}(x,y),\mathsf{WildAnimal}(y)\rightarrow\mathsf{ZooClass}(x)$, assuming that $\mathsf{Image},  \mathsf{WildAnimal}\in\mathsf{CN}$, $\mathsf{depicts}\in\mathsf{RN}$, and $\mathsf{ZooClass}\in{\mathcal{C}}$. It is important that explanation rules refer only to individuals $a \in\mathsf{EN}$ that correspond to items $\mathcal{M}(a) \in \mathcal{D}$; this is guaranteed by the conjunct $\mathsf{Exemplar}(x)$ in the explanation rule body. Indeed, since the classifier under explanation is unknown, the only guaranteed information is the classification of the exemplars. 

Given a classifier $F:\mathcal{D}\rightarrow{\mathcal{C}}$ and a set of individuals $\mathcal{I}\subseteq{\mathsf{EN}}$, the positive set (pos-set) of $F$ on $\mathcal{I}$ for class $C\in{\mathcal{C}}$ is  $\mathsf{pos}(F,\mathcal{I},C)=\{a\in{\mathcal{I}}:F(\mathcal{M}(a))=C\}$. 

\begin{definition}[Explanation Rule Correctness] \label{def:explanation rule correctness}

Let $F:\mathcal{D}\rightarrow{\mathcal{C}}$ be a classifier, $\mathcal{E}=\langle{ \mathcal{M},\mathcal{S}}\rangle$ an explanation dataset in terms of $\mathcal{D}$, $\mathcal{C}$ and an appropriate vocabulary $\mathcal{V}$, and $\rho(F,\mathcal{E},C)$ an explanation rule. 
The rule $\rho$ is \emph{correct} if and only if
\begin{align*}
    \mathsf{fol}(\mathcal{S}&\cup\{\mathsf{Exemplar}\sqsubseteq \{a\mid a\in\mathsf{EN}\}\}\\&\phantom{xxxxx}\cup\{C(a)\mid a\in \mathsf{pos}(F,\mathsf{EN},C)\})\models \rho
\end{align*}
where 
$\mathsf{fol}(\mathcal{K})$ is the first-order logic translation of DL KB $\mathcal{K}$.
\end{definition}

The intended meaning of a correct explanation rule is that for every $a \in \mathsf{EN}$, if the body of the rule holds, then the classifier classifies $\mathcal{M}(a)$ to the class indicated in the head of the rule.
Intuitively, an explanation rule is correct if it is a logical consequence of the underlying knowledge extended by the axiom $\mathsf{Exemplar}\sqsubseteq \{a\mid a\in\mathsf{EN}\}$, which forces $\mathsf{Exemplar}(x)$ be true in an interpretation $\mathcal{I}$ only for $x=a^\mathcal{I}$ with $a\in\mathsf{EN}$.

For instance, the rule of the previous example $\mathsf{Exemplar}(x),\mathsf{Image}(x),\mathsf{depicts}(x,y),\mathsf{WildAnimal}(y)\rightarrow \mathsf{ZooClass}(x)$ would be correct for the KB $\mathcal{S}_1=\langle \mathcal{T}_1, \mathcal{A}_1\rangle$, where $\mathcal{A}_1=\{\mathsf{Image}(\mathsf{a}), \mathsf{depicts}(\mathsf{a},\mathsf{b}),\mathsf{Wolf}(\mathsf{b})\}$ and $\mathcal{T}_1=\{\mathsf{Wolf\sqsubseteq{WildAnimal}}\}$ if $\mathsf{a}\in{\mathsf{pos}(F,\mathsf{EN},\mathsf{ZooClass})}$, while it would not be correct for the KB $\mathcal{S}_2=\langle \emptyset,\mathcal{A}_1\rangle$, nor would it be correct for $\mathcal{S}_1$ if $\mathsf{a}\not\in{\mathsf{pos}(F,\mathsf{EN},\mathsf{ZooClass})}$.  
Checking whether a rule is correct is a reasoning problem which can be solved by using standard DL reasoners. On the other hand, finding rules which are correct is an inverse problem which is much harder to solve.

\section{Computing explanation rules from queries }\label{sec:Semantic queries as explanation rules}

As mentioned in Sec. \ref{sec:preliminaries}, an instance query has the form  $\{c_1,\ldots, c_n\}_x$, which resembles the body of an explanation rule with head some $C(x)$. Thus, by representing the bodies of explanation rules as queries,  the computation of explanations can be treated as a query reverse engineering problem. 

\begin{definition}[Explanation Rule Query] \label{def:explantion rule query}

Let $F:\mathcal{D}\rightarrow{\mathcal{C}}$ be a classifier, $\mathcal{E}=\langle{ \mathcal{M},\mathcal{S}}\rangle$ an explanation dataset in terms of $\mathcal{D}$, $\mathcal{C}$ and an appropriate vocabulary $\mathcal{V}$, and $\rho(F,\mathcal{E},C)$: $\mathsf{Exemplar}(x), c_1, c_2,\dots, c_n\rightarrow{C(x)}$ an explanation rule. 
The instance query
\[
q_\rho \doteq\{\mathsf{Exemplar}(x),c_1,c_2,\dots,c_n \}_x
\]
\noindent is the \emph{explanation rule query} of explanation rule $\rho$.
\end{definition}

This definition establishes a 1-1 relation (up to variable renaming) between $\rho$ and $q_\rho$. 
To compute queries corresponding to explanation rules 
that are guaranteed to be correct, we prove Theorem \ref{thrm:semantic queries and rules}.

\begin{theorem}\label{thrm:semantic queries and rules}

Let $F:\mathcal{D}\rightarrow{\mathcal{C}}$ be a classifier, $\mathcal{E}=\langle{ \mathcal{M},\mathcal{S}}\rangle$ an explanation dataset in terms of $\mathcal{D}$, $\mathcal{C}$ and an appropriate vocabulary $\mathcal{V}$, $\rho(F,\mathcal{E},C)$: $\mathsf{Exemplar}(x),c_1,c_2,\dots,c_n\rightarrow C(x)$ an explanation rule,   
and $q_\rho$ the explanation rule query of $\rho$. The explanation rule $\rho$ is correct if and only if

\[
\mathsf{cert}(q_\rho,\mathcal{S})\subseteq \mathsf{pos}(F,\mathsf{EN},C) 
\]

\end{theorem}

Theorem~\ref{thrm:semantic queries and rules} (see appendix for proof) allows us to compute guaranteed correct rules, by finding a query $q$ for which $\mathsf{cert}(q,\mathcal{S})\subseteq{\mathsf{pos}(F,\mathsf{EN},C)}$. Intuitively, an explanation rule query is correct for class $C$, if all of its certain answers are mapped by $\mathcal{M}$ to feature data which is classified in class $C$. It follows that a query with  one certain answer which is an element of the pos-set is a correct rule query, as is a query $q$ for which $\mathsf{cert}(q,\mathcal{S})=\mathsf{pos}(F,\mathsf{EN},C)$. Thus, it is useful to define a \emph{recall} metric for explanation rule queries by comparing the set of certain answers with the pos-set of a class~$C$: 
\[\mathsf{recall}(q,\mathcal{E},C)=\frac{|\mathsf{cert}(q,\mathcal{S})\cap\mathsf{pos}(F,\mathsf{EN},C)|}{|\mathsf{pos}(F,\mathsf{EN},C)|}.\]

Given the above, one approach to the problem of finding explanation rules for an explanation dataset is to reduce it to forming candidate queries, computing their answers, and assessing the correctness of the corresponding rules.

The computation of arbitrary candidate explanation rule queries for the KB $\mathcal{S}$ of an explanation dataset is in general hard since it involves exploring the query space $\mathcal{Q}$ of all queries that can be constructed using the underlying vocabulary $\mathcal{V}$ and getting their certain answers for $\mathcal{S}$.
Difficulties arise even in simple cases, since the query space is in general infinite. 
However, the set of all possible distinct answer sets is finite and in most cases it is expected to be much smaller than its upper limit, the powerset $2^{\mathsf{IN}}$.

Alg.~\ref{alg-qsg} explores 
a useful finite subset of $\mathcal{Q}$, namely the tree-shaped queries of a  maximum depth $k$ \cite{DBLP:conf/ijcai/GlimmHLS07}. It constructs all possible such queries (that include $\mathsf{Exemplar}(x)$  in the body), obtains their answers, and arranges them in a directed acyclic graph (the \emph{query space DAG}) using the subset relation on the answer sets. The queries are constructed in the for loop, and then the while loop replaces  queries having the same answer set by their intersection. The \emph{intersection} $q_1\sqcap q_2$ of two instance queries $q_1$, $q_2$ with answer variable $x$ is the query $\mathsf{cond}(q_1\cup q_2\theta)$, where $\theta$ renames each variable appearing in $q_2$ apart from $x$ to a variable not appearing in $q_1$. Thus, from all possible queries with the same answers, the algorithm keeps only the \emph{most specific} query $q$ of all such queries. Intuitively, this is the most detailed query.  
Finally, the queries are arranged in a DAG. By construction, each node of the DAG is a query representing a distinct answer set. 

\begin{figure}[t]\vspace*{-10pt}
\begin{algorithm}[H]
\SetAlgoLined
\KwData{Vocabulary $\mathcal{V}$, KB $\mathcal{K}$, a maximum query depth $k\geq0$}
\KwResult{Query space DAG $\mathcal{G}$ }

Compute the set $\mathcal{B}$ of all non-syntactically equivalent queries 
$\{C_1(x),\ldots, C_n(x)\}_x$, where $C_i\in\mathsf{CN}\kern-1pt\setminus\kern-2pt\{\mathsf{Exemplar}\}$, $n\kern-0.7pt\geq\kern-0.7pt 1$.\;

Compute the set $\mathcal{F}$ of all non-syntactically equivalent queries  $\{r_1(u_{1},v_{1}),\ldots, r_n(u_{n},v_{n})\}_{x,y}$, where $r_i\in\mathsf{RN}$, $n\geq1$, each $u_{i}$, $v_{i}$ is either $x$ or $y$ and $u_{i}\neq v_{i}$.\;

Initialize an empty set of queries $\mathcal{Q}$.\;

\For{$i=0\ldots k$}{
Compute the set $\mathcal{T}_i$ of all trees of depth $i$.\;

\ForEach{$t\in \mathcal{T}_i$}{
Assign to each node $v$ of $t$ a distinct variable $\mathsf{var}(v)$. Assign $x$ to the root of $t$.\;

Construct all non-syntactically equivalent queries $q$ obtained from $t$ by adding to the body of $q$: i) for each node $v$ of $t$, the body of an element of $\mathcal{B}\cup\{\emptyset\}$ after renaming $x$ to $\mathsf{var}(v)$, ii) for each edge $(v_1,v_2)$ of $t$, the body of an element of $\mathcal{F}$ after renaming $x$ to $\mathsf{var}(v_1)$ and $y$ to $\mathsf{var}(v_2)$, and iii) $\mathsf{Exemplar}(x)$.\;

Condense all $q$s and add them to $\mathcal{Q}$.\;
}
}
\While{\rm there are $q_1,q_2\in\mathcal{Q}$ s.t.\! $\mathsf{cert}(q_1,\mathcal{K})=\mathsf{cert}(q_2,\mathcal{K})$}{remove $q_1,q_2$ from $\mathcal{Q}$ and add $q_1\sqcap q_2$ to $\mathcal{Q}$.\;
}

Arrange the elements of $\mathcal{Q}$ in a DAG $\mathcal{G}$, making $q_1$ a child of $q_2$ iff $\mathsf{cert}(q_1,\mathcal{K})\subset\mathsf{cert}(q_2,\mathcal{K})$.\;

\Return{{\rm the transitive reduction of} $\mathcal{G}$} 
\caption{QuerySpaceDAG}\label{alg-qsg}
\end{algorithm}\vspace*{-16pt}
\end{figure}

\begin{theorem}\label{lemma-1}
Let $F:\mathcal{D}\rightarrow{\mathcal{C}}$ be a classifier, $\mathcal{E}=\langle{ \mathcal{M},\mathcal{S}}\rangle$ an explanation dataset in terms of $\mathcal{D}$, $\mathcal{C}$ and an appropriate vocabulary $\mathcal{V}$, and $\rho(F,\mathcal{E},C)$ a correct tree-shaped explanation rule of maximum depth~$k$.
The DAG constructed by Alg.~\ref{alg-qsg} contains a query $q_{\rho'}$
corresponding to a correct explanation rule  $\rho'(F,\mathcal{E},C)$ with the same metrics as $\rho$, s.t. $q_{\rho'}\leq_S q_\rho$. 
\end{theorem}

Given Theorem~\ref{lemma-1} (see appendix for proof) 
a node corresponding to a correct rule for some $\mathsf{pos}(F,\mathsf{EN},C)$ can be reached by traversing the graph starting from the root and finding the first node whose answer set equals $\mathsf{pos}(F,\mathsf{EN},C)$. The descendants of that node provide all queries  corresponding to correct explanation rules. The DAG has a unique root because answer sets are subsets of $\mathsf{cert}(\{\mathsf{Exemplar}(x)\}_x,\mathcal{S})$.

An unavoidable difficulty in using Alg.~\ref{alg-qsg} is its complexity. The sizes of $\mathcal{B}$ and $\mathcal{F}$ are at the orders of $2^{|\mathsf{CN}|}$ and $4^{|\mathsf{RN}|}$ respectively, and the number of tree-shaped queries with  $k$ variables is at the order of $2^{k|\mathsf{CN}|}\cdot 4^{(k-1)|\mathsf{RN}|}$. However, in practice the query space is much smaller since most queries have zero answers and can be ignored. To get answer sets, Alg.~\ref{alg-qsg} assumes a function that returns $\mathsf{cert}(q,\mathcal{K})$ for any query $q$. 

If $\mathcal{K}$ is fully materialized, i.e. if no reasoning is needed to answer queries, it is easy to implement the function for $\mathsf{cert}(q,\mathcal{K})$. The sets $\mathcal{B}$ and $\mathcal{F}$ can be computed to contain only queries with at least one answer, and queries can be constructed incrementally; once a query with no answers is reached, no queries with additional conjuncts are considered. 

If $\mathcal{K}$ is not materialized, or impossible to materialize, the incremental query construction process should by coupled with the necessary reasoning to get the query answers. For the DL-Lite$^-_{\mathcal{R}}$ dialect, a more efficient alternative to Alg.~\ref{alg-qsg} is proposed in \cite{DBLP:conf/dlog/ChortarasGS19}. DL-Lite$^-_{\mathcal{R}}$  
allows only axioms of the form $C\sqsubseteq D$ or $r\sqsubseteq s$, where $C,D$ are concepts, and $r,s$ are atomic roles. $D$ can be either atomic or of the form $\exists r^{(-)}.\top$, and $C$ can be of the form $\exists r^{(-)}.A$, where $A$ is atomic. The authors exploit the fact that query answering in DL-Lite$^-_{\mathcal{R}}$ can be done in steps by rewriting a query to a set of queries, the union of whose answer sets are the answers to the original query, to incrementally compute the  tree-shaped queries of a maximum depth with at least one answer.

Further simplifications to reduce the practical complexity of Alg.~\ref{alg-qsg}, that may affect its theoretical properties, include not condensing queries, keeping an arbitrary query for each answer set instead of the most specific one, and setting a minimum answer set size threshold for a query to be considered.

\section{Experiments and Evaluation}\label{sec-evaluation}

In this section we evaluate 
the proposed approach, which we call KGrules. 
We conduct experiments on tabular and image data, investigating how explanation datasets of different sizes and expressivities affect the quality of the explanations, we compare our work with other rule-based explanation methods, and discuss the quality and usability of the results.
Regarding Alg.~\ref{alg-qsg}, we implemented its  adaptation for DL-Lite$^-_{\mathcal{R}}$ described in \cite{DBLP:conf/dlog/ChortarasGS19} with the additional simplifications mentioned above. 

\subsection{Tabular Classifier}
\label{subsec:exp-tabular}
The first set of experiments is conducted on the Mushroom\footnote{https://archive.ics.uci.edu/ml/datasets/mushroom} dataset which contains data in tabular form with categorical features. 
 Our proposed approach is overkill for such a dataset, since its representation as a DL KB does not contain roles nor a TBox, however on this dataset we can compare  the proposed method with the state-of-the-art. 
To represent the dataset as a knowledge base in order to run Alg.\ref{alg-qsg}, we create a concept for each combination of categorical feature name and value, ending up with $|\mathsf{CN}|=123$ and an individual for each row of the dataset. Then we construct an ABox where the type of each individual is asserted based on the values of its features and the aforementioned concepts. 
 To measure the quality of the generated explanations, the dataset is split in three parts: A classification-training set on which we train a simple two layer MLP classifier, an explanation-training set which we use to generate explanations for the predictions of the classifier with the methods under evaluation, and an explanation-testing set on which we measure the fidelity (ratio of input instances on which the predictions of the model and the rules agree, over total instances) of the rules.
We also measure the number of rules and the average rule length for each case.
We compare our method with Skope-Rules\footnote{https://github.com/scikit-learn-contrib/Skope-Rules}
and rule-matrix \cite{DBLP:journals/tvcg/MingQB19} which implements scalable bayesian rule lists \cite{DBLP:conf/icml/YangRS17}, 
on different sizes of explanation-training sets. The results are shown in Table~\ref{tab:mushroom}. All methods perform similarly with respect to fidelity, with no clear superiority of any method since they all achieve near perfect performance, 
probably because of the simplicity of the dataset.    

\subsection{Image Classifier: CLEVR-Hans3}
Nowadays the explanation of 
tabular classifiers is not a main concern, and the experiments of Sec. \ref{subsec:exp-tabular} show that such tasks are rather easy. Of 
interest is the explanation of deep learning models that take as input raw data, like images or text.
Hence, for the second set of experiments, we employ CLEVR-Hans3 \cite{stammer2020right} which is a dataset of images with intentionally added biases in the train and validation set which are absent in the test set. For example the characteristic of the first class is that all images include a large cube and a large cylinder, but the large cube is always gray in the training and validation sets, while it has a random color in the test set. This makes it ideal for the evaluation of XAI 
frameworks since it creates classifiers with foreknown biases. 
On this dataset we conduct two experiments. First we explore the effect of the size of the explanation dataset by testing whether our method can 
predict the known description of the three classes. 
Secondly, we evaluate our  
method on a real image classifier and compare the results with other rule-based methods.

For representing available annotations as a DL KB, we define an individual name for each image and for each object depicted therein, and a concept name for each color, size, shape and material of the objects. We also include a role name $\mathsf{contains}$, to connect images to objects they depict. Then, in the ABox, we assert the characteristics of each object and link them to the appropriate images by using the role. For comparing with other methods, we also create a tabular version of the dataset, in which each object's characteristics are one-hot encoded, and an image is represented as a concatenation of the encodings of the objects it depicts.

\begin{table}[t]
  \small

    \begin{tabular}{c c c c c}
      \toprule 
      \textbf{Size} & \textbf{Method} & \textbf{Fidelity} & \textbf{Nr. of Rules}& \textbf{Avg. Length} \\
      \midrule 
      \multirow{3}{*}{\rotatebox{90}{100}}
      & KGrules 
      & \textbf{97.56\%}
      & 11
      & 5 \\
      & RuleMatrix 
      & 94.53\%
      & 3
      & 2 \\
      & Skope-Rules 
      & 97.01\%
      & 3
      & 2 \\
      \hline
      \multirow{3}{*}{\rotatebox{90}{{200}}} 
      & KGrules 
      & 98.37\%
      & 11
      & 5 \\
      & RuleMatrix 
      & 97.78\%
      & 4
      & 2 \\
      & Skope-Rules 
      & \textbf{98.49\%}
      & 4
      & 2 \\
      \hline
      \multirow{3}{*}{\rotatebox{90}{{600}}} 
      & KGrules 
      & 99.41\%
      & 13
      & 4 \\
      & RuleMatrix 
      & \textbf{99.43\%}
      & 6
      & 1 \\
      & Skope-Rules 
      & 98.52\%
      & 4
      & 2 \\
      \bottomrule 
    \end{tabular}
  \caption{Performance 
  on the Mushroom dataset.}\label{tab:mushroom}\vspace*{-3pt}
\end{table}

Using the true labels of the data allows us to use the description of each class as ground truth explanations. With explanation datasets with 600 or more exemplars we are able to predict 
the ground truth for all 3 classes. Even with 20 exemplars we are able to produce the correct explanation for one of the classes and with 40 or more exemplars we produce ground truth explanations for 2 out of the 3 classes and almost the correct explanation for the third class (only one characteristic of one object missing). 
In order to produce accurate explanations it seems useful to have individuals 
close to the ``semantic border" of the classes, i.e. individuals
of different classes with similar descriptions. Intuitively, such individuals guide the algorithm to produce a more accurate explanation in a similar manner that near-border examples guide a machine learning algorithm to approximate better the separating function. Following this intuition, we experiment with two of the small explanation datasets that almost found the perfect explanations (size of 40 and 80). By strategically choosing individuals
, we are able to obtain two small explanation datasets, one of size 43 and one of size 82, that when used by Alg.~\ref{alg-qsg} produce the correct explanations for all 3 classes. This indicates the importance of the curation of the explanation dataset, which is not an easy task,  
and the selection of ``good" individuals for the explanation dataset is not trivial. 

Finally we use our framework to explain a real classifier trained on CLEVR-Hans3, and compare our explanations with Skope-Rules and RuleMatrix. The classifier we use is a ResNet34-based model, that achieved overall 99,4\% validation accuracy and 71,2\% test accuracy (probably due to the confounded train and validation sets). More details about the classifier's performance 
can be found in the technical appendix. We curate an explanation set with 100 images so that it also accurately explains the ground truth, and we used it to explain the real classifier.  Table~\ref{tab:clevr-compare} shows that our method significantly outperforms the other rule-based methods in terms of fidelity, with a notable smaller number of rules which is used as an indication of understandability of a rule-set. The set nature of the input data (each image contains a set of objects with specific characteristics) shows the limitations of other rule-based methods in such realistic problems. We are able to reproduce the rules created by our method using the tabular format that is fed to the other classifiers, showing that the data format is not a limitation in terms of fidelity, but it requires a large number of rules, which indicates the usefulness of rule-based methods like ours that don't only work on tabular data. Investigating the explanations produced by our method, we are also able to detect potential biases of the classifier due to the confounding factors of the dataset. For example, regarding the first class (all images contain a large cube and a large cylinder), the rule with the highest recall produced for the real classifier is: $\mathsf{contains}(x,y),$ $\mathsf{Gray}(y),$ $\mathsf{Large}(y),$ $\mathsf{contains}(x,z),$ $\mathsf{Cylinder}(z),$ $\mathsf{Large(z)} \rightarrow \mathsf{Class}_1(x)$ showing the existence of a large cylinder, and detecting the potential color bias of another large object created by the intentional bias of the train and validation set (the large cube is always gray in the train and validation sets).   

\begin{table}[t]
  \small
    \begin{tabular}{c c c c c}
      \toprule 
      \textbf{Method} & \textbf{Fidelity} & \textbf{Nr. of Rules}& \textbf{Avg. Length} \\
      \midrule 
      KGrules 
      & \textbf{85.07\%}
      & 4
      & 5 \\
      RuleMatrix 
      & 58.09\%
      & 42
      & 2 \\
      Skope-Rules 
      & 77.18\%
      & 20
      & 3 \\
      \bottomrule 
    \end{tabular}
  \caption{Performance 
  on the CLEVR-Hans3 dataset.}\label{tab:clevr-compare}\vspace*{-3pt}
\end{table}

\subsection{Image Classifier: Visual Genome}
As a third experiment, we evaluate our framework on an explanation dataset of real-world images, described by an expressive knowledge, that includes roles and a TBox.
Specifically, we utilize 
the Visual Genome dataset (VGD)  \cite{DBLP:journals/ijcv/KrishnaZGJHKCKL17} which contains richly annotated images, including descriptions of regions, attributes of depicted objects and relations between them. On this dataset, we attempt to explain image classifiers trained on ImageNet. We define three super-classes of ImageNet classes which contain a) Domestic, b) Wild and c) Aquatic Animals
, because they are more intuitive to perform a qualitative evaluation, when compared to the fine-grained ImageNet classes.
We represent the available VGD annotations as a DL KB, where the ABox consists of the scene graphs for each image, in which each node and edge is labeled with a WordNet (WN) synset and the TBox consists of the WN hypernym-hyponym hierarchy. In the ABox we also include assertions about which objects are depicted by an image in order to connect the exemplar data with the scene graphs. Since in the original VGD annotations are linked to WN automatically, there are errors, thus we chose to manually curate a subset of 100 images. This is closer to the intended use-case of our proposed method, in which experts would curate explanation datasets for specific domains. 
We explain three different neural 
architectures\footnote{https://pytorch.org/vision/stable/models.html}: VGG-16, Wide-ResNet (WRN) and ResNeXt, trained for  
classification on the ImageNet dataset. The context of VGD is too complex to be transformed into tabular form in a useful and valid way for the other rule-based methods. 
Table~\ref{tab:VG-explanations} shows the correct rules of maximum recall for each class and each classifier. 
We discuss three key explanations:

\begin{table}[t]
    \begin{center}
      {\small
\begin{tabular}{c l}
      \toprule 
      \textbf{Net.} & \textbf{Rules}  \\
      \midrule 
      \multirow{3}{*}{\rotatebox{90}{VGG-16}} 
      & $\mathsf{artifact}(y),\mathsf{dog}(z),\mathsf{brown}(w) \rightarrow \mathsf{Domestic}(x)$ \\ 
      & $\mathsf{green}(y),\mathsf{plant}(z),\mathsf{organ}(w) \rightarrow \mathsf{Wild}(x)$ \\
      & $\mathsf{whole}(y),\mathsf{ocean}(z) \rightarrow \mathsf{Aquatic}(x)$ \\           \midrule
      \multirow{3}{*}{\rotatebox{90}{WRN}} 
      & $\mathsf{animal}(y),\mathsf{wear}(y,z),\mathsf{artifact}(z) \rightarrow \mathsf{Domestic}(x)$ \\
      & $\mathsf{green}(y),\mathsf{plant}(z),\mathsf{nose}(w) \rightarrow \mathsf{Wild}(x)$ \\
      & $\mathsf{surfboard}(y) \rightarrow \mathsf{Aquatic}(x)$ \\ 
      \midrule
      \multirow{3}{*}{\rotatebox{90}{ResNext}} 
      & $\mathsf{artifact}(y),\mathsf{dog}(z),\mathsf{brown}(w) \rightarrow \mathsf{Domestic}(x)$ \\ 
      & $\mathsf{ear}(y),\mathsf{plant}(z),\mathsf{nose}(w) \rightarrow \mathsf{Wild}(x)$ \\
      & $\mathsf{fish}(y), \mathsf{structure}(z) \rightarrow \mathsf{Aquatic}(x)$ \\ 
      \bottomrule 
    \end{tabular}}%
  \end{center}\vspace*{-5pt}
 \caption{Explanation rules utilizing the animal explanation dataset. Rules are shown in condensed form: the full rules are obtained by adding the conjuncts $\mathsf{contains}(x,t)$ for all appearing variables $x\kern-1pt\neq \kern-1pt t$. }\label{tab:VG-explanations}
\end{table}

1. Wide ResNet: $\mathsf{surfboard}(y) \rightarrow$ \emph{Aquatic(x)}. It seems that the classifier has a bias accepting surfer/surfboard images as aquatic animals probably due to the sea environment of the images; further investigation finds this claim to be consistent, showing the potential of this framework in detecting biases. 

2. Wide ResNet: $\mathsf{animal}(y)$, $\mathsf{wear}(y,z)$, $\mathsf{artifact}(z)$ $\rightarrow$ \emph{Domestic(x)}. It is interesting to compare this explanation with another correct rule for the same classifier 
 with lower recall: $\mathsf{animal}(y)$, $\mathsf{collar}(z)$ $\rightarrow$ \emph{Domestic(x)}. By considering roles between objects we get a more accurate (higher recall) and informative explanation, denoting the tendency of the classifier to classify as \emph{Domestic} any animal that wears something man-made. This example shows how more complex queries enhance the insight (wearing an artifact) while less expressive ones might only see a part of it (collar). Here we can also see one of the effects of the TBox hierarchy on the explanations, since this rule covers many sub-cases (like dog wears collar, and cat wears bowtie) that would require multiple rules if it wasn't for the grouping that stems from the TBox. 

3. ResNeXt: $\mathsf{nose}(y)$, $\mathsf{plant}(z)$, $\mathsf{ear}(w)$ $\rightarrow$ \emph{Wild}. Although this explanation provides information that is  
related to the nature environment of the images classified as \emph{Wild} (plant), we see also some rather odd concepts (nose, ear). While this could be a strange bias of the classifier, it is probably a flaw of the explanation dataset. As we discovered, images are not consistently annotated with body parts, like noses and ears. Thus, through the explanations we can also detect weaknesses of the explanation set. The rules
are limited by the available knowledge, so we should constantly evaluate the quality and expressivity of the knowledge that is used in order to produce accurate and useful explanations.

\section{Conclusions}\label{sec-conc}
In this work, we introduced a framework for representing explanations for ML classifiers in the form of rules, developed an algorithm for computing KG-based explanations, generated explanations for various classifiers and datasets, and compared our work with other methods. We believe that the transparency of the proposed explanation dataset, combined with the guarantees of framework and algorithm improve \emph{user awareness} when compared with other rule-based explanation methods. As with comparing understandability however, this should be evaluated in a human study, which we plan to conduct in the future. In addition, we are in the process of creating explanation datasets in collaboration with domain experts for the domains of medicine and music. We are investigating what constitutes a “good” explanation dataset with regards to its size, distribution and represented information. Finally, we are exploring improvements and optimizations for the algorithm, its adaptation to more DL dialects, relaxations for getting approximate solutions faster, and modifications in order to generate different types of explanations such as local, counterfactual, and prototype or example based explanations.
\bibliographystyle{named}
\bibliography{ijcai22}

\appendix

\section{Proofs}

\subsection{Proof of Theorem~1}
 \begin{proof}
Let $\mathcal{S}=\langle\mathcal{T},\mathcal{A}\rangle$. Because by definition $\mathsf{Exemplar}(a)\in \mathcal{A}$ iff $a\in\mathsf{EN}$ and $\mathsf{Exemplar}$ does not appear anywhere in $\mathcal{T}$, we have
 \begin{align*}
 \mathsf{cert}(q_\rho,\mathcal{S})&=\mathsf{cert}(\{\mathsf{Exemplar},c_1,\ldots, c_n\},\langle\mathcal{T},\mathcal{A}\rangle)\\
 &\hskip-30pt=\mathsf{EN}\cap \mathsf{cert}(\{c_1,\ldots, c_n\},\langle\mathcal{T},\mathcal{A}\rangle)\\
 &\hskip-30pt=\mathsf{cert}(\{\mathsf{Exemplar}\},\\
 &\hskip10pt\langle\{\mathsf{Exemplar}\sqsubseteq \{a\mid a\in\mathsf{EN}\}\},\mathcal{A}\rangle)\,\cap\\ &\hskip17pt\mathsf{cert}(\{c_1,\ldots, c_n\},\langle\mathcal{T},\mathcal{A}\rangle)\\
 &\hskip-30pt=\mathsf{cert}(\{\mathsf{Exemplar},c_1,\ldots, c_n\},\\
 &\hskip10pt\langle\mathcal{T}\cup\{\mathsf{Exemplar}\sqsubseteq \{a\mid a\in\mathsf{EN}\}\},\mathcal{A})\\
 &\hskip-30pt=\mathsf{cert}(q_\rho,\langle\mathcal{T}\cup\{\mathsf{Exemplar}\sqsubseteq \{a\mid a\in\mathsf{EN}\}\},\mathcal{A})
 \end{align*}


Because by definition $C$ does not appear anywhere in $\mathcal{S})$, we have also that 
 $\mathsf{cert}(q_\rho,\mathcal{S})= \mathsf{cert}(q_\rho,\mathcal{S'})$, where $\mathcal{S'}=\mathcal{S}\cup\{\mathsf{Exemplar}\sqsubseteq \{a\mid a\in\mathsf{EN}\}\}\cup\{C(a)\mid a\in \mathsf{pos}(F,\mathsf{EN},C)\}\}$, since  the assertions $C(a)$ are not involved neither in the query nor in $\mathcal{S}$ and hence have no effect.

 By definition of a certain answer, $e\in\mathsf{cert}(q,\mathcal{K})$ iff for every model $\mathcal{I}$ of $\mathcal{K}$ there is a match $\pi$ s.t. $\pi(x)=e^\mathcal{I}$ and $\pi(u)\in D^\mathcal{I}$ for all $D(u)\in q$ and $(\pi(u),\pi(v))\in r^\mathcal{I}$ for all $r(u,v)\in q$.

 Assume that $\rho$ is correct and let $e\in\mathsf{cert}(q_\rho,\mathcal{S})$. We have proved that also  $e\in\mathsf{cert}(q_\rho,\mathcal{S}')$.  Because $\rho$ is correct, by Def.~3 it follows that  every model $\mathcal{I}$ of $\mathcal{S}'$ is also a model of $\rho$. Because the body of $q_\rho$ is the same as the body of $\rho$, $\pi$ makes true both the body of $\rho$ and the head of $\rho$, which is $C(x)$, hence $e^\mathcal{I}\in C^\mathcal{I}$. It follows that $C(e)$ is true in $\mathcal{I}$. But the only assertions of the form $C(e)$ in $\mathcal{S}'$ are the assertions $\{C(a) \mid a\in \mathsf{pos}(F,\mathsf{EN},C)\}$, thus $e\in \mathsf{pos}(F,\mathsf{EN},C)$.

 For the inverse, assume that $\mathsf{cert}(q_\rho,\mathcal{S})\subseteq \mathsf{pos}(F,\mathsf{EN},C)$, equivalently  $\mathsf{cert}(q_\rho,\mathcal{S'})\subseteq \mathsf{pos}(F,\mathsf{EN},C)$. Thus if $e\in\mathsf{cert}(q_\rho,\mathcal{S})$ then $e^\mathcal{I}\in C^\mathcal{I}$. Since this holds for every model $\mathcal{I}$ of $\mathcal{S}'$ and the body of $q_\rho$ is the same as the body of $\rho$, it follows that $\mathcal{I}$ is also a model of $\rho$, i.e. $\rho$ is correct.

 \end{proof}

\subsection{Proof of Theorem~2}

\begin{proof}
Because $\rho$ is correct,  $\mathsf{cert}(q_\rho,\mathcal{S})\subseteq\mathsf{pos}(F,\mathsf{EN},C)$. We know that $q_\rho$ is in the set of all tree-shaped queries of maximum depth $k$, hence there is a node $q$ in the DAG such that $\mathsf{cert}(q,\mathcal{S})=\mathsf{cert}(q_\rho,\mathcal{S})$ with the property that $q\leq_S q'$ for all $q'$ such that $\mathsf{cert}(q,\mathcal{S})=\mathsf{cert}(q',\mathcal{S})$, because $q$ is the most specific query with that answer set (by construction, $q\doteq\mathsf{cond}(\bigsqcap q')$). It follows that $q_\rho$ must be one of those $q'$s, hence  $q\leq_S q_\rho$. Because $\mathsf{cert}(q,\mathcal{S})=\mathsf{cert}(q_\rho,\mathcal{S})$, if $\rho'$ is the explanation rule corresponding to $q$, then obviously $\rho$ and $\rho'$ have the recall, precision etc. metrics.\\
\end{proof}

\section{ResNet-34 Performance on CLEVR-Hans3}

\begin{table}[h]
\centering
\caption{Test Set Metrics of ResNet-34 on CLEVR-Hans3.}
\label{table:chmetrics}
\begin{tabular}{cccc}
\toprule
True label & Precision & Recall & F1-score\\
\midrule
Class 1 & 0.94 & 0.16 & 0.27 \\
Class 2 & 0.59 & 0.98 & 0.54 \\
Class 3 & 0.85 & 1.00 & 0.92\\
\bottomrule
\end{tabular}
\end{table}

\begin{table}[h]
\centering
\caption{Confusion Matrix of ResNet-34 on CLEVR-Hans3.}
\label{table:chmetrics}
\begin{tabular}{c c c c c c}
\toprule
& &\multicolumn{3}{c}{Predicted} \\
\cmidrule{3-5}
& & Class 1 & Class 2 & Class 3\\
\cmidrule{2-5}
\parbox[t]{2mm}{\multirow{3}{*}{\rotatebox[origin=c]{90}{True}}} &
Class 1 & 118 & 511 & 121 \\
& Class 2 & 5 & 736 & 9 \\
& Class 3 & 2 & 0 & 748\\
\bottomrule
\end{tabular}
\end{table}

\end{document}